\documentclass[conference]{IEEEtran}

\usepackage{cite}
\usepackage{amsmath,amssymb,amsfonts}
\usepackage{graphicx}
\usepackage{booktabs}
\usepackage{multirow}
\usepackage{url}
\usepackage{xcolor}
\usepackage{array}
\usepackage{algorithmic}
\usepackage{textcomp}
\usepackage{listings}
\usepackage{enumitem}
\usepackage{tikz}
\usetikzlibrary{positioning,arrows.meta,shapes.geometric}
\tikzset{
block/.style={
    draw,
    rounded corners,
    thick,
    align=center,
    minimum width=3.2cm,
    minimum height=0.9cm,
    fill=blue!6
},
greenblock/.style={
    draw,
    rounded corners,
    thick,
    align=center,
    minimum width=3.2cm,
    minimum height=0.9cm,
    fill=green!10
},
orangeblock/.style={
    draw,
    rounded corners,
    thick,
    align=center,
    minimum width=3.2cm,
    minimum height=0.9cm,
    fill=orange!15
},
redblock/.style={
    draw,
    rounded corners,
    thick,
    align=center,
    minimum width=3.2cm,
    minimum height=0.9cm,
    fill=red!10
},
arrow/.style={
    -{Stealth},
    thick
},
dashedarrow/.style={
    -{Stealth},
    thick,
    dashed
}
}

\title{FST.ai 2.5: Explainable and Uncertainty-Aware AI for Olympic and Para-Taekwondo Decision Support, Athlete Digital Twins, and Federation-Scale Analytics}

\author{
\IEEEauthorblockN{
Keivan Shariatmadar$^{1}$,
Ahmad Osman$^{1}$,
Ramin Rey$^{2}$
}

\IEEEauthorblockA{
$^{1}$AutomaTiQ Research Group,
htw saar University of Applied Sciences and Fraunhofer IZFP,
Saarbrücken, Germany\\
\{keivan.shariatmadar, ahmad.osman\}@htwsaar.de
}

\IEEEauthorblockA{
$^{2}$Sport Director, Austrian Taekwondo Union (ATU)
}
}

\begin{document}

\maketitle

\begin{abstract}

The rapid digitalisation of elite sport has created unprecedented opportunities for the integration of artificial intelligence, performance analytics, and decision-support systems into athlete development and competition management. However, existing solutions remain fragmented, often focusing on isolated tasks such as performance analysis, athlete monitoring, or referee support. This paper presents \textbf{FST$\cdot$ai 2.5}, an explainable, uncertainty-aware, and secure artificial intelligence framework designed for Olympic and Para-Taekwondo applications. 
The proposed framework introduces a unified digital ecosystem that integrates athlete intelligence, competition analytics, federation-scale data management, AI-assisted decision support, simulation environments, and future referee-support capabilities. Central to the framework are athlete and event digital twins, explainable performance indicators, adaptive training recommendations, and secure federation-aware governance mechanisms that support World Taekwondo (WT), Member National Associations (MNAs), coaches, referees, analysts, and athletes.

FST$\cdot$ai 2.5 combines multi-source competition evidence, athlete-performance data, and contextual information to generate actionable insights through explainable AI models and uncertainty-aware analytics. The framework supports tactical diagnostics, longitudinal athlete monitoring, performance forecasting, personalised development planning, and federation-wide benchmarking while preserving transparency, traceability, and data security. Prototype deployments and pilot federation-scale infrastructures are presented to demonstrate the feasibility of the proposed approach. 
Although developed for Olympic and Para-Taekwondo, the underlying methodology provides a general foundation for explainable AI, digital-twin technologies, and uncertainty-aware decision support in combat sports and other high-performance sporting environments.

\end{abstract}


\section{Introduction}

Artificial intelligence (AI) is rapidly transforming modern sport by enabling data-driven decision making, automated performance analysis, athlete monitoring, tactical assessment, and competition analytics. Recent advances in computer vision, deep learning, sensor technologies, and explainable artificial intelligence (XAI) have created new opportunities for supporting athletes, coaches, referees, and sporting organisations through intelligent decision-support systems \cite{Claudino2019,Arrieta2020,Fernandez2021}. These technologies are increasingly being adopted across elite sports to improve performance evaluation, talent identification, injury-risk management, tactical preparation, and long-term athlete development. 
Combat sports represent particularly challenging environments for AI applications. Unlike many team sports, decision making in combat sports occurs within highly dynamic and rapidly changing environments characterised by short reaction times, complex tactical interactions, incomplete information, and substantial uncertainty. In Olympic Taekwondo, athletes continuously adapt their behaviour in response to their opponents, match context, score evolution, referee decisions, and strategic objectives. Simultaneously, referees must interpret high-speed actions under significant time constraints while maintaining fairness, consistency, and transparency. These characteristics make Taekwondo a demanding but highly relevant domain for the development of advanced AI-based decision-support systems. 
The digital transformation of Taekwondo has accelerated considerably during the last decade. Modern competitions employ electronic Protector and Scoring Systems (PSS), video replay technologies, digital competition-management platforms, and large-scale event databases. Consequently, substantial amounts of information are generated during every competition, including scoring events, penalties, referee actions, timing information, athlete statistics, video recordings, and tactical event histories. Despite the availability of these data sources, their utilisation remains largely fragmented. Existing systems typically focus on isolated functionalities such as scoring, event management, video replay, or statistical reporting. As a result, valuable information is often distributed across disconnected platforms and is rarely transformed into actionable athlete intelligence or federation-level decision support. 
Recent research has demonstrated promising applications of computer vision and machine learning in sports performance analysis, activity recognition, pose estimation, and tactical assessment \cite{Claudino2019,Fernandez2021}. Furthermore, digital-twin technologies have emerged as powerful tools for representing complex physical systems through continuously updated virtual counterparts \cite{Barricelli2020,Fuller2020,Kritzinger2018}. Explainable AI has similarly gained increasing attention as a mechanism for improving transparency, trustworthiness, and user acceptance of AI-generated recommendations \cite{Arrieta2020,Guidotti2018}. However, most existing approaches focus on individual components rather than providing an integrated ecosystem capable of supporting athletes, coaches, referees, federations, and governing bodies simultaneously. 
An additional challenge arises from the presence of uncertainty throughout the athlete-development process. Competition outcomes, tactical behaviour, athlete readiness, injury risk, and performance evolution are influenced by incomplete observations, measurement limitations, contextual factors, and human decision making. Traditional AI systems frequently provide point estimates without explicitly communicating uncertainty, thereby limiting their usefulness in high-stakes sporting environments. Recent developments in uncertainty-aware AI and Epistemic AI suggest that incorporating uncertainty information can substantially improve robustness, interpretability, and decision quality \cite{Shariatmadar2024Uncertainty,Wang2024Credal}. 
To address these limitations, this paper presents \textbf{FST$\cdot$ai 2.5}, an explainable, uncertainty-aware, and secure artificial intelligence ecosystem for Olympic and Para-Taekwondo. Rather than focusing exclusively on athlete performance analysis or action recognition, the proposed framework introduces a federation-scale digital ecosystem that integrates athlete intelligence, competition analytics, referee support, simulation capabilities, explainable decision support, and secure governance mechanisms within a unified architecture. 
The proposed framework introduces several key capabilities: 
\begin{itemize}
    \item Explainable AI-assisted referee-support and education modules;
    \item Athlete intelligence, readiness assessment, and KPI analytics;
    \item Athlete and event digital twins for longitudinal performance modelling;
    \item AI-based tactical diagnostics and performance forecasting;
    \item Simulation-driven athlete evaluation and scenario analysis;
    \item Adaptive AI-generated training recommendations;
    \item Dedicated Olympic and Para-Taekwondo support services;
    \item Secure federation-scale infrastructure with role-based access control;
    \item Future real-time video-analysis and decision-support services;
    \item Uncertainty-aware analytics based on Epistemic AI principles.
\end{itemize}
The framework is designed to support World Taekwondo (WT), Member National Associations (MNAs), coaches, referees, analysts, sport scientists, and athletes within a unified and secure operational environment. By combining explainable AI, digital twins, uncertainty-aware analytics, and federation-scale governance, FST$\cdot$ai 2.5 establishes a foundation for next-generation intelligent decision-support systems in combat sports. 
The remainder of this paper is organised as follows. Section II reviews related work in sports analytics, digital twins, explainable AI, and uncertainty-aware decision support. Section III presents the overall FST$\cdot$ai 2.5 architecture and system components. Sections IV–VII describe athlete intelligence, digital-twin modelling, federation-scale analytics, and AI-assisted decision-support services. Finally, the paper discusses deployment considerations, future research directions, and opportunities for broader adoption across combat sports and elite sporting environments.

\section{Related Work}
The application of artificial intelligence in sport has expanded significantly over the last decade, driven by advances in machine learning, computer vision, wearable sensing, and data analytics. Modern sports organisations increasingly rely on AI-driven technologies for performance monitoring, injury-risk assessment, tactical analysis, talent identification, and decision support \cite{Claudino2019,Fernandez2021,VanEetvelde2021}. These developments have transformed large volumes of competition and training data into valuable resources for evidence-based athlete development and strategic decision making.

\subsection{AI and Sports Analytics}
Sports analytics has evolved from descriptive statistical reporting toward predictive and prescriptive decision-support systems. Machine-learning methods have been applied to performance forecasting, injury prediction, workload monitoring, and athlete profiling across a wide range of sports \cite{Claudino2019,Halson2014,Gabbett2016,Windt2017}. Recent studies have demonstrated the potential of AI systems to support coaches and sport scientists by identifying performance patterns that are difficult to detect through traditional analysis alone. Nevertheless, many existing solutions remain focused on isolated analytical tasks and rarely provide integrated environments capable of supporting athletes, coaches, federations, and governing bodies simultaneously.

\subsection{Computer Vision and Human Action Recognition}
Computer vision has become one of the most active research areas in sports AI. Advances in pose estimation, object detection, and activity recognition have enabled automated analysis of athlete movement and tactical behaviour. Deep-learning architectures based on convolutional neural networks, graph neural networks, and transformer-based models have demonstrated strong performance in human action recognition and video understanding tasks \cite{Cao2021,Dosovitskiy2021}. These approaches provide the technological foundation for automated event detection, technical analysis, referee support, and athlete-performance assessment. However, most systems focus primarily on recognition accuracy and provide limited interpretability for end users.

\subsection{Digital Twins in Sport}
Digital-twin technologies have emerged as a powerful paradigm for representing complex physical systems through continuously updated virtual counterparts \cite{Barricelli2020,Fuller2020,Kritzinger2018}. While digital twins have been widely investigated in manufacturing, healthcare, and industrial systems, their adoption in sport remains relatively limited. Recent research suggests that athlete-centred digital twins can support longitudinal performance monitoring, simulation, decision support, and personalised development planning. Nevertheless, comprehensive frameworks integrating athlete digital twins, event digital twins, competition analytics, and federation-level decision support remain largely unexplored.

\subsection{Explainable and Uncertainty-Aware Artificial Intelligence}
The growing adoption of AI in high-stakes environments has highlighted the importance of explainability, transparency, and trustworthiness. Explainable Artificial Intelligence (XAI) seeks to provide interpretable reasoning processes that enable users to understand and validate AI-generated recommendations \cite{Arrieta2020,Guidotti2018}. In parallel, recent developments in uncertainty-aware AI and Epistemic AI have demonstrated the importance of distinguishing between reliable and uncertain predictions, particularly in environments characterised by incomplete information and dynamic decision making \cite{Shariatmadar2024Uncertainty,Wang2024Credal}. These concepts are especially relevant in sport, where athlete performance, competition outcomes, and tactical decisions are inherently uncertain.

\subsection{Artificial Intelligence in Taekwondo}
Research on AI applications in Taekwondo has primarily focused on biomechanical analysis, motion recognition, scoring evaluation, wearable sensing, and athlete-performance assessment \cite{Bridge2009,Kazemi2006,Tornello2014,Shariatmadar2025FST}. Although these studies have demonstrated the feasibility of AI-assisted analysis in combat sports, most existing systems remain limited to specific analytical tasks and do not provide federation-scale infrastructures capable of supporting athlete intelligence, referee development, coaching support, competition analytics, and governance within a unified ecosystem.

\subsection{Research Gap and Contribution}
Despite substantial progress in sports analytics, several important limitations remain. Existing solutions frequently suffer from limited explainability, insufficient treatment of uncertainty, fragmented analytical workflows, a lack of federation-scale deployment capabilities, minimal support for Para-Taekwondo, and weak integration between competition analytics and athlete-development processes. Furthermore, few systems provide secure multi-tenant infrastructures capable of supporting athletes, coaches, referees, Member National Associations (MNAs), and governing bodies within a common platform. 
FST$\cdot$ai 2.5 addresses these limitations by introducing an integrated framework that combines athlete and event digital twins, explainable and uncertainty-aware AI, federation-scale analytics, secure governance mechanisms, referee-support services, simulation environments, and adaptive athlete-development tools. The framework, therefore, extends beyond traditional performance-analysis systems and establishes a foundation for next-generation intelligent decision-support ecosystems in Olympic and Para-Taekwondo.

\section{FST$\cdot$ai 2.5 Framework}
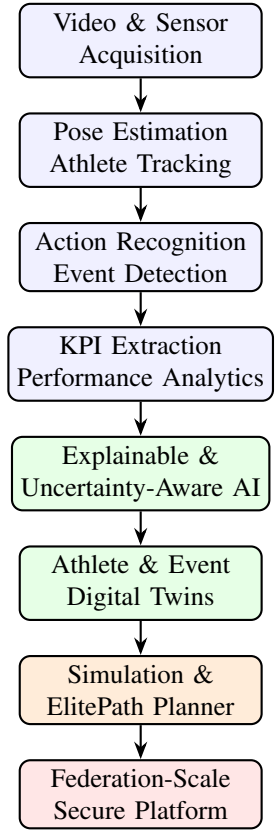
\begin{figure}[ht]
\centering
\begin{tikzpicture}[node distance=0.45cm]
\node[block] (video) {Video \& Sensor\\Acquisition};

\node[block,below=of video] (pose) {Pose Estimation\\Athlete Tracking};

\node[block,below=of pose] (action) {Action Recognition\\Event Detection};

\node[block,below=of action] (kpi) {KPI Extraction\\Performance Analytics};

\node[greenblock,below=of kpi] (xai) {Explainable \&\\Uncertainty-Aware AI};

\node[greenblock,below=of xai] (twin) {Athlete \& Event\\Digital Twins};

\node[orangeblock,below=of twin] (planner) {Simulation \&\\ElitePath Planner};

\node[redblock,below=of planner] (fed) {Federation-Scale\\Secure Platform};

\draw[arrow] (video)--(pose);
\draw[arrow] (pose)--(action);
\draw[arrow] (action)--(kpi);
\draw[arrow] (kpi)--(xai);
\draw[arrow] (xai)--(twin);
\draw[arrow] (twin)--(planner);
\draw[arrow] (planner)--(fed);
\end{tikzpicture}

\caption{High-level architecture of FST$\cdot$ai 2.5. Competition evidence is progressively transformed into explainable athlete intelligence, digital twins, and federation-scale decision-support services.}
\label{fig:fst_architecture}
\end{figure}

FST$\cdot$ai 2.5 is a next-generation artificial intelligence ecosystem designed to support Olympic and Para-Taekwondo through explainable analytics, uncertainty-aware reasoning, athlete intelligence, referee-support services, and federation-scale decision-support capabilities. The framework extends traditional sports analytics systems by integrating competition analysis, athlete development, digital twins, simulation environments, and secure governance mechanisms within a unified operational platform. 
The development of FST$\cdot$ai 2.5 was motivated by several practical challenges observed in modern Taekwondo. While contemporary competitions generate substantial amounts of information through electronic scoring systems, video recordings, athlete databases, and competition-management platforms, these data sources are typically fragmented and rarely transformed into actionable intelligence. Coaches often rely on manual video review, referees receive limited objective feedback, and federations lack integrated tools capable of supporting longitudinal athlete development, performance monitoring, and strategic decision making. 
The proposed framework addresses these limitations by establishing a complete intelligence pipeline that transforms raw competition evidence into explainable and actionable knowledge. Video streams, scoring information, athlete profiles, physiological assessments, and competition histories are integrated to construct continuously evolving athlete and event digital twins. These digital representations enable advanced performance analysis, tactical diagnostics, forecasting, simulation, and adaptive development planning while maintaining transparency and traceability throughout the decision-making process. 
Unlike conventional AI systems that focus primarily on prediction accuracy, FST$\cdot$ai 2.5 adopts a human-centred design philosophy. The framework emphasises explainability, uncertainty awareness, and operational usability to ensure that generated recommendations can be understood, validated, and trusted by athletes, coaches, referees, analysts, and federation officials. This is particularly important in sporting environments, where AI systems are intended to support human expertise rather than replace it. 
Furthermore, the framework is designed to operate at multiple organisational levels. At the athlete level, it supports performance analysis, readiness estimation, and personalised development planning. At the coaching level, it provides tactical insights and evidence-based training recommendations. At the federation level, it enables athlete benchmarking, talent-development monitoring, and strategic performance evaluation. Finally, at the governing-body level, the framework establishes the foundation for future referee-support systems, competition analytics, educational services, and large-scale sport-intelligence infrastructures. 
The overall architecture of FST$\cdot$ai 2.5 is illustrated in Figure~\ref{fig:fst_architecture}. The framework consists of several interconnected layers responsible for data acquisition, athlete tracking, action recognition, performance analytics, explainable reasoning, digital-twin construction, simulation, planning, and federation-scale governance. Together, these components form a comprehensive AI ecosystem for supporting evidence-based decision making throughout the athlete lifecycle.

\subsection{Overall Architecture}
FST$\cdot$ai 2.5 is designed as a modular, explainable, and uncertainty-aware artificial intelligence ecosystem for Olympic and Para-Taekwondo. The framework integrates athlete-performance analytics, competition intelligence, referee-support services, digital twins, simulation environments, and federation-scale management within a unified architecture. 
The system follows a multi-layer design in which raw competition evidence is progressively transformed into actionable decision-support information. Unlike conventional sports-analysis tools that focus on isolated tasks such as pose estimation or match statistics, FST$\cdot$ai 2.5 establishes an end-to-end intelligence pipeline capable of supporting athletes, coaches, referees, analysts, Member National Associations (MNAs), and World Taekwondo (WT) through a common operational platform. 
The architecture consists of eight interconnected layers:
\begin{enumerate}
    \item Multi-camera video acquisition and competition-data collection;
    \item Athlete detection, tracking, and pose estimation;
    \item Action recognition and tactical event understanding;
    \item KPI extraction and athlete-performance analytics;
    \item Explainable and uncertainty-aware reasoning;
    \item Athlete and event digital twins;
    \item AI-assisted planning, simulation, and forecasting;
    \item Federation-scale secure infrastructure and governance.
\end{enumerate}
The modular design enables the integration of future technologies, including electronic Protector and Scoring Systems (PSS), wearable sensors, physiological monitoring devices, computer-vision services, and real-time officiating-support modules.

\subsection{Action Recognition and Pose Estimation}
Automated understanding of athlete behaviour constitutes one of the foundational capabilities of FST$\cdot$ai 2.5. The framework employs computer-vision pipelines that detect athletes, estimate body landmarks, reconstruct motion trajectories, and identify technical and tactical actions from video streams. 
Modern pose-estimation frameworks such as OpenPose and transformer-based vision architectures provide detailed skeletal representations of athlete motion, enabling the extraction of biomechanical and tactical information from competition footage \cite{Cao2021,Dosovitskiy2021}. These representations form the basis for higher-level activity recognition and tactical analysis. 
The framework supports the identification of:
\begin{itemize}
    \item body kicks,
    \item head kicks,
    \item turning kicks,
    \item spinning techniques,
    \item punches,
    \item clinch situations,
    \item boundary violations,
    \item movement trajectories,
    \item reaction times,
    \item tactical transitions.
\end{itemize}
Future developments will investigate graph neural networks, transformer-based temporal models, and multimodal architectures capable of combining video, scoring-system information, and wearable-sensor data within a unified representation.

\subsection{KPI Extraction and Athlete Intelligence}
Following action recognition, the detected events are transformed into performance indicators that describe athlete behaviour, technical execution, tactical effectiveness, and competitive performance. 
Examples of extracted KPIs include: 
\begin{itemize}
    \item scoring efficiency,
    \item attack volume,
    \item defensive effectiveness,
    \item technical diversity,
    \item reaction-time indicators,
    \item penalty frequency,
    \item round-specific performance,
    \item tactical consistency,
    \item success rates of specific techniques.
\end{itemize}
These indicators are aggregated across matches and competitions to create longitudinal athlete profiles that support athlete intelligence, benchmarking, progression monitoring, and performance forecasting.

\subsection{Explainability and Uncertainty-Aware Reasoning}
Because FST$\cdot$ai 2.5 is intended for high-stakes sporting environments, explainability represents a fundamental design requirement. Decisions affecting athlete development, referee education, competition analysis, and federation management must be transparent and scientifically justifiable. 
The framework, therefore, incorporates explainable AI mechanisms that provide interpretable reasoning pathways, confidence estimates, and evidence-based explanations for generated recommendations \cite{Arrieta2020,Guidotti2018}. In addition, uncertainty-aware analytics derived from Epistemic AI principles enable the system to distinguish between reliable conclusions and predictions affected by incomplete information \cite{Shariatmadar2024Uncertainty,Wang2024Credal}. 
The explainability layer supports:
\begin{itemize}
    \item confidence estimation,
    \item uncertainty quantification,
    \item tactical reasoning summaries,
    \item visual evidence overlays,
    \item decision traceability,
    \item explainable KPI generation.
\end{itemize}
These capabilities improve user trust and facilitate practical adoption by coaches, referees, athletes, and federation officials.

\section{AI Decision-Support Framework}
\label{sec:decision_support}
The primary objective of FST$\cdot$ai 2.5 is not merely the collection and visualisation of competition data but the generation of actionable decision support for athletes, coaches, referees, analysts, Member National Associations (MNAs), and governing bodies. Modern Taekwondo environments generate large volumes of heterogeneous information, including competition outcomes, scoring events, tactical patterns, physiological assessments, training records, and video evidence. Transforming this information into meaningful recommendations requires a systematic decision-support framework capable of integrating evidence, quantifying uncertainty, and providing transparent recommendations. 
Unlike conventional sports analytics platforms that primarily deliver descriptive statistics, FST$\cdot$ai 2.5 is designed as an explainable decision-support ecosystem. The framework combines athlete intelligence, tactical analysis, simulation, uncertainty quantification, and human expertise to assist decision makers while maintaining human authority over final actions.

\subsection{Decision Hierarchy}
Decision making within elite sport occurs at multiple organisational levels. FST$\cdot$ai 2.5, therefore, supports three complementary categories of decisions. 
\textbf{Operational decisions} correspond to short-term actions performed during training sessions or competitions. Examples include tactical adaptations, match preparation, referee assessments, and opponent-specific recommendations. 
\textbf{Tactical decisions} involve medium-term planning activities such as training adaptation, competition scheduling, athlete selection, and performance-development strategies. 
\textbf{Strategic decisions} focus on long-term objectives, including athlete development pathways, talent identification, federation planning, resource allocation, and international performance benchmarking. 
Let
\[
\mathcal{D}
=
\{
d_1,d_2,\ldots,d_n
\}
\]
denote the set of candidate decisions and
\[
\mathcal{E}
=
\{
e_1,e_2,\ldots,e_m
\}
\]
the available evidence extracted from competitions, diagnostics, athlete records, and analytical services. 
The objective of the decision-support engine is to identify the recommendation that maximises expected utility:
\[
d^*
=
\arg\max_{d\in\mathcal{D}}
U(d\mid\mathcal{E}),
\]
where \(U(\cdot)\) denotes the utility associated with a particular decision given the available evidence.
This formulation establishes a direct relationship between collected evidence and generated recommendations while maintaining flexibility across different decision-making contexts.

\subsection{Explainable Recommendation Generation}
A fundamental requirement for adoption within sport-governance environments is explainability. Coaches, referees, and federation officials must understand not only the recommendation itself but also the underlying evidence supporting that recommendation. 
For each recommendation \(d\), the platform generates an explanation tuple
\[
\mathcal{X}(d)
=
\{
\mathcal{E}(d),
C(d),
R(d)
\},
\]
where
\begin{itemize}
\item \(\mathcal{E}(d)\) denotes the supporting evidence,
\item \(C(d)\) denotes the confidence associated with the recommendation,
\item \(R(d)\) denotes the reasoning process used by the system.
\end{itemize}
Consequently, recommendations become transparent and auditable rather than opaque outputs of a black-box model. For example, a recommendation to increase head-kick training intensity may be supported by observed deficiencies in scoring conversion rates, low tactical diversity, and declining late-round performance.

\subsection{Human-in-the-Loop Decision Support}
FST$\cdot$ai 2.5 is based on the principle that artificial intelligence should augment rather than replace human expertise. Coaches and referees possess contextual understanding, psychological insight, and experiential knowledge that cannot be fully represented by analytical models alone. 
The final decision can therefore be expressed as
\[
D_{\text{final}}
=
\lambda_h D_h
+
(1-\lambda_h)D_{AI},
\]
where
\begin{itemize}
\item \(D_h\) denotes the human decision,
\item \(D_{AI}\) denotes the AI-generated recommendation,
\item \(\lambda_h\in[0,1]\) controls the influence of human judgement.
\end{itemize}
In practice, FST$\cdot$ai 2.5 operates as an intelligent assistant rather than an autonomous decision maker. The framework provides evidence, explanations, and recommendations, while final authority remains with the responsible human stakeholders.

\subsection{Uncertainty-Aware Decision Support}
Athlete performance and competition outcomes are inherently uncertain. Missing data, incomplete observations, noisy measurements, and limited historical evidence may affect the reliability of generated recommendations. 
To address this challenge, FST$\cdot$ai 2.5 explicitly models uncertainty. Let
\[
U_i
=
w_v U_v
+
w_h U_h
+
w_d U_d
+
w_c U_c,
\]
where
\begin{itemize}
\item \(U_v\) represents uncertainty originating from video analysis,
\item \(U_h\) represents uncertainty associated with historical data,
\item \(U_d\) represents uncertainty arising from diagnostic measurements,
\item \(U_c\) represents contextual uncertainty.
\end{itemize}
The weighting coefficients satisfy
\[
w_v+w_h+w_d+w_c=1.
\]
Recommendations are therefore accompanied by confidence estimates that allow users to distinguish between conclusions supported by strong evidence and those requiring additional observations.

\subsection{Simulation-Based Decision Support}
In addition to descriptive analytics, FST$\cdot$ai 2.5 supports simulation-based reasoning. Rather than analysing only the current athlete state, the framework can estimate the consequences of hypothetical interventions. 
Let
\[
\mathbf{s}_i(t)
\]
denote the current athlete's state and
\[
\mathbf{u}
\]
a candidate intervention. The simulated future state is
\[
\mathbf{s}_i(t+\tau)
=
f
\big(
\mathbf{s}_i(t),
\mathbf{u}
\big),
\]
where \(f(\cdot)\) represents the simulation model. 
The objective is to estimate the expected improvement
\[
\Delta R_i
=
R_i(t+\tau)-R_i(t),
\]
where \(R_i\) denotes athlete readiness. 
This capability enables coaches and federations to evaluate alternative training strategies before implementation and therefore supports more evidence-based athlete development.
\begin{figure*}[ht!]
\centering
\begin{tikzpicture}[node distance=.7cm]

\node[block] (evidence) {Competition Evidence\\Diagnostics\\Video Analytics};

\node[block,below=of evidence] (fusion)
{Evidence Fusion};

\node[block,below=of fusion] (analytics)
{AI Analytics};

\node[block,below left=1cm and 1.5cm of analytics] (forecast)
{Forecasting};

\node[block,below=of analytics] (uncertainty)
{Uncertainty\\Estimation};

\node[block,below right=1cm and 1.5cm of analytics] (simulation)
{Simulation};

\node[greenblock,below=1cm of uncertainty] (decision)
{Decision-Support Engine};

\node[orangeblock,below=of decision] (human)
{Coach / Referee / Federation};

\draw[arrow] (evidence)--(fusion);
\draw[arrow] (fusion)--(analytics);

\draw[arrow] (analytics)--(forecast);
\draw[arrow] (analytics)--(uncertainty);
\draw[arrow] (analytics)--(simulation);

\draw[arrow] (forecast)--(decision);
\draw[arrow] (uncertainty)--(decision);
\draw[arrow] (simulation)--(decision);

\draw[arrow] (decision)--(human);

\end{tikzpicture}
\caption{
Decision-support architecture implemented within FST$\cdot$ai 2.5. Competition evidence, diagnostics, and video analytics are transformed into forecasts, uncertainty estimates, and simulation results that support human-centred decision making.
}
\label{fig:decision_support}
\end{figure*}
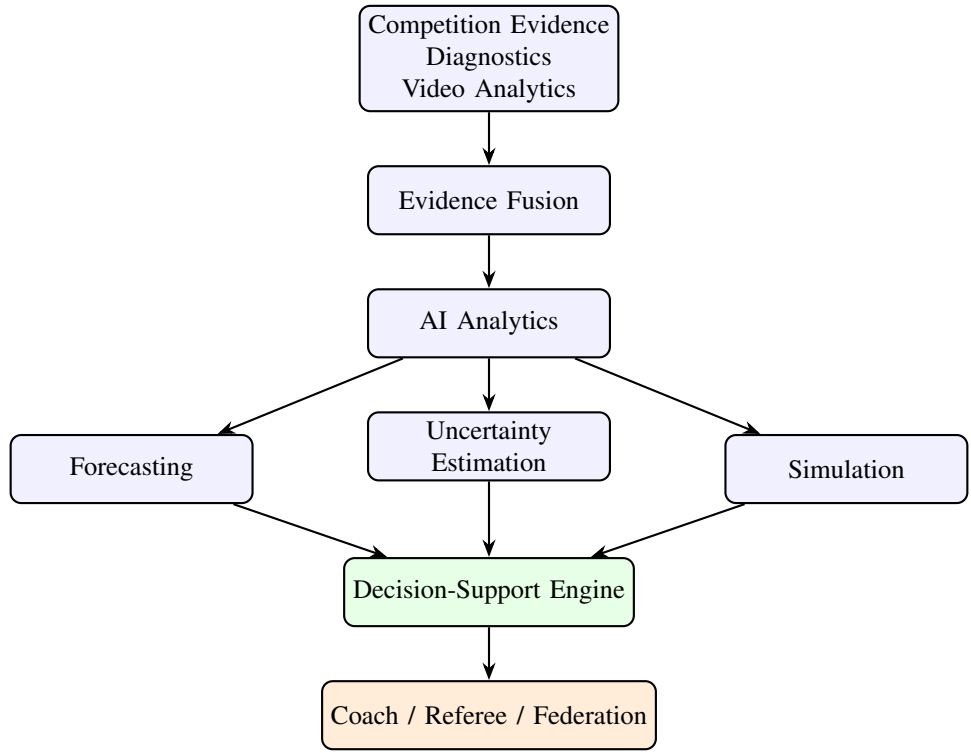

\section{Explainable AI for Referee and Coaching Support}
\label{sec:xai}
Trust and transparency represent fundamental requirements for AI adoption within sport-governance environments. Unlike many commercial AI systems that operate as black-box predictors, decision-support systems used by referees, coaches, and federations must provide understandable reasoning processes capable of explaining why a recommendation was generated. 
Recent developments in Explainable Artificial Intelligence (XAI) emphasise the importance of transparency, interpretability, and accountability in high-stakes decision-making environments \cite{Arrieta2020,Guidotti2018}. FST$\cdot$ai 2.5, therefore incorporates explainability as a core architectural component rather than an optional add-on.

\subsection{Explanation Model}
Each recommendation generated by the framework is associated with an explanation structure
\[
\mathcal X
=
\{
\mathcal E,
C,
R
\},
\]
where
\begin{itemize}
\item \(\mathcal E\) denotes supporting evidence,
\item \(C\) denotes confidence,
\item \(R\) denotes reasoning.
\end{itemize}
This representation allows users to understand not only the recommendation itself but also the evidence and logic supporting it.

\subsection{Referee Support}
Future versions of FST$\cdot$ai are expected to support referees through automated event detection, confidence estimation, and explainable recommendation services. 
Let
\[
V
\]
denote a video segment and
\[
A(V)
\]
the detected action. 
The referee-support system generates
\[
\hat y
=
g(A(V)),
\]
where \(\hat y\) represents a candidate interpretation of the observed event. 
Rather than automatically enforcing decisions, the system provides:
\begin{itemize}
\item detected action,
\item confidence score,
\item supporting evidence,
\item rule references,
\item explanation summary.
\end{itemize}
The final authority remains with the referee.

\subsection{Coaching Support} 
For coaching applications, explainability is equally important. Recommendations regarding athlete development, tactical adaptation, and training planning must be understandable and actionable. 
Consequently, FST$\cdot$ai generates recommendations together with supporting evidence such as:
\begin{itemize}
\item scoring efficiency,
\item attack volume,
\item tactical diversity,
\item endurance indicators,
\item readiness estimates,
\item uncertainty measures.
\end{itemize}
This evidence-based approach improves trust while facilitating practical adoption within real coaching environments.
\begin{figure}[t]
\centering
\begin{tikzpicture}[node distance=.5cm]

\node[block] (video) {Video / Match Data};

\node[block,below=of video]
(action)
{Action Recognition};

\node[block,below=of action]
(pred)
{Prediction};

\node[block,below=of pred]
(conf)
{Confidence};

\node[block,below=of conf]
(explain)
{Explanation Layer};

\node[greenblock,below=of explain]
(user)
{Coach / Referee};

\draw[arrow] (video)--(action);
\draw[arrow] (action)--(pred);
\draw[arrow] (pred)--(conf);
\draw[arrow] (conf)--(explain);
\draw[arrow] (explain)--(user);

\end{tikzpicture}
\caption{Explainable-AI workflow for coaching and referee support.}
\label{fig:xai}
\end{figure}
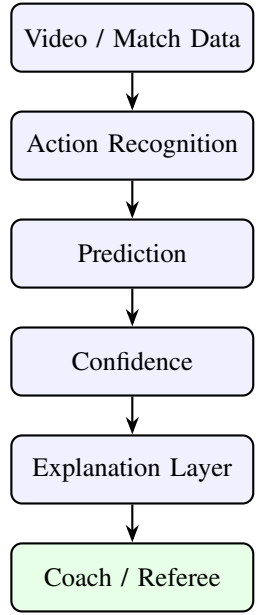

\section{Secure Federation-Scale Athlete Intelligence Platform}
\label{sec:secure_portal}
\begin{figure*}[t]
\centering
\begin{tikzpicture}[node distance=.5cm]

\node[redblock] (wt) {World Taekwondo};

\node[block,below left=1cm and 2cm of wt] (mna) {Member National\\Associations};

\node[block,below right=1cm and 2cm of wt] (coach) {Coaches / Analysts};

\node[greenblock,below=1cm of wt] (platform)
{FST$\cdot$ai Secure Athlete Intelligence Platform};

\node[block,below left=1cm and 2cm of platform] (db)
{Federation-Aware\\Database};

\node[block,below=1cm of platform] (ai)
{AI Analytics\\Engine};

\node[block,below right=1cm and 2cm of platform] (sim)
{Simulation \&\\ElitePath};

\node[orangeblock,below=1cm of ai] (twin)
{Athlete Digital Twins};

\draw[arrow] (wt)--(mna);
\draw[arrow] (wt)--(platform);
\draw[arrow] (wt)--(coach);

\draw[arrow] (platform)--(db);
\draw[arrow] (platform)--(ai);
\draw[arrow] (platform)--(sim);

\draw[arrow] (db)--(ai);
\draw[arrow] (ai)--(sim);
\draw[arrow] (sim)--(platform);
\draw[arrow] (ai)--(twin);

\end{tikzpicture}
\caption{Conceptual architecture of the secure federation-scale athlete intelligence platform within FST$\cdot$ai 2.5. The platform integrates federation governance, athlete digital twins, AI analytics, simulation services, and adaptive planning within a secure multi-tenant environment.}
\label{fig:secure_platform}
\end{figure*}
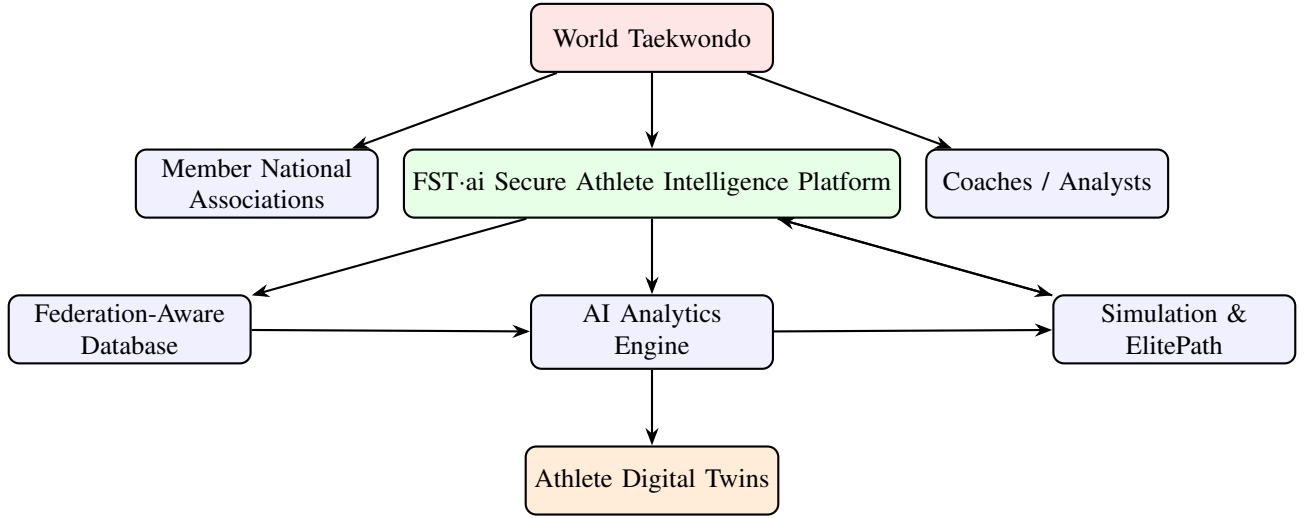

The increasing digitalisation of elite sport has created unprecedented opportunities for athlete monitoring, performance analytics, competition intelligence, and evidence-based decision making. However, the growing volume of athlete-related information also introduces significant challenges concerning data management, privacy, governance, interoperability, and secure access control. Modern sporting ecosystems involve multiple stakeholders, including athletes, coaches, referees, analysts, national federations, and international governing bodies, each requiring different levels of access to sensitive information. 
Traditional athlete-management systems typically focus on administrative record keeping and competition results. While such systems provide useful storage capabilities, they often lack advanced analytical services, secure federation-level governance mechanisms, explainable decision-support tools, and integrated athlete-development functionality. Furthermore, data are frequently fragmented across independent databases, spreadsheets, competition-management platforms, video repositories, and coaching systems, making comprehensive athlete intelligence difficult to achieve. 
To address these challenges, FST$\cdot$ai 2.5 introduces a secure federation-scale athlete intelligence platform that extends the analytical capabilities of the framework into a deployable operational ecosystem. The platform combines athlete management, KPI analytics, digital twins, simulation services, adaptive planning, and governance mechanisms within a unified infrastructure. Particular emphasis is placed on federation-aware data isolation, role-based access control, privacy preservation, and explainable AI services to ensure that analytical outputs can be deployed safely within real sporting environments. 
The proposed platform is designed to support multiple organisational layers simultaneously. At the athlete level, it provides personalised performance analytics and development recommendations. At the coaching level, it delivers evidence-based insights and training support tools. At the federation level, it enables athlete benchmarking, talent-development monitoring, and strategic performance management. At the international level, it establishes the technological foundation for future competition intelligence, referee-support services, educational systems, and global athlete-development initiatives. 
Figure~\ref{fig:secure_platform} illustrates the high-level architecture of the secure federation-scale athlete intelligence platform. The architecture integrates secure authentication services, federation-aware databases, athlete digital twins, AI analytics engines, simulation services, and adaptive planning modules into a common ecosystem capable of supporting large-scale deployment across multiple federations and governing bodies.

\subsection{Motivation and Vision}
Beyond AI-assisted decision support and performance analysis, FST$\cdot$ai 2.5 introduces a secure federation-scale athlete intelligence platform designed for World Taekwondo (WT), Member National Associations (MNAs), coaches, analysts, referees, and athletes. The objective is to move beyond isolated analytical modules and establish a unified digital infrastructure for securely storing, managing, analysing, and interpreting athlete-related data across different organisational levels. 
The platform is motivated by a practical challenge in international sport: athlete data are distributed across federations, event systems, coaching records, diagnostic assessments, and competition reports. Without a secure and federation-aware architecture, such information remains fragmented and difficult to transform into reliable athlete intelligence. FST$\cdot$ai 2.5 addresses this issue by combining role-based access control, federation-level data isolation, athlete-performance analytics \cite{GDPR2016,Simmhan2005}, simulation services, and AI-supported training recommendations within a single operational ecosystem. 
The proposed infrastructure, therefore, extends FST$\cdot$ai from an analytical framework into a deployable digital environment capable of supporting athlete development, competition analytics, tactical preparation, Para-Taekwondo support, referee education, and future real-time decision-support services.

\subsection{System Architecture}
The athlete intelligence platform follows a multi-layered architecture composed of a user-facing interface, secure backend services, an AI analytics layer, and a federation-aware database infrastructure. Each layer is designed to support modularity, scalability, and secure deployment. 
At the prototype level, the platform is implemented using FastAPI backend services, a PostgreSQL relational database, JSON Web Token (JWT)-based authentication, Docker-based deployment, and federation-level role isolation. Although these technologies represent one implementation pathway, the conceptual architecture is technology-independent and can be migrated to cloud-native or enterprise infrastructures. 
The architecture consists of five principal components:
\begin{enumerate}
    \item \textbf{Frontend Interface:} provides dashboards for WT administrators, MNA users, coaches, analysts, and athletes.
    \item \textbf{Secure Backend Services:} manage authentication, authorisation, API access, data validation, and business logic.
    \item \textbf{Federation-Aware Database:} stores athlete profiles, KPI records, simulation outputs, training recommendations, and audit trails.
    \item \textbf{AI Analytics Layer:} computes athlete-performance indicators, readiness estimates, simulation outputs, and training priorities.
    \item \textbf{Governance and Security Layer:} enforces access control, federation-level isolation, auditability, and privacy constraints.
\end{enumerate}

\subsection{Federation-Level Role Hierarchy}
A central design requirement of FST$\cdot$ai 2.5 is strict separation between different organisational roles. The platform therefore implements a hierarchical role model supporting:
\begin{itemize}
    \item \texttt{WT\_ADMIN},
    \item \texttt{MNA\_ADMIN},
    \item \texttt{COACH},
    \item \texttt{ANALYST},
    \item \texttt{ATHLETE}.
\end{itemize}
Each user role is associated with different permissions. WT administrators can manage global configurations, federations, and system-level data. MNA administrators manage athletes and coaches within their federation. Coaches access athlete profiles, KPI dashboards, and training recommendations for authorised athletes. Analysts access performance and simulation outputs, while athletes may view selected personal analytics and development plans. 
Federation-level isolation is achieved by associating athlete records, KPI entries, simulation outputs, and training recommendations with a federation identifier. This ensures that users can only access data within their authorised organisational scope. Such a design is essential for international sport environments, where multiple federations may participate in the same platform while requiring strict data separation.

\subsection{Athlete Management and KPI Infrastructure}
The platform supports structured athlete-profile management, longitudinal KPI tracking, tactical statistics, and event-based performance analysis. Athlete records may include demographic information, sport classification, competition history, training status, performance indicators, Para-Taekwondo category information, and development objectives. 
Stored KPI indicators include:
\begin{itemize}
    \item head-kick accuracy,
    \item body-kick efficiency,
    \item defensive reaction time,
    \item endurance and late-round performance indices,
    \item clinch-control indicators,
    \item boundary-risk behaviour,
    \item penalty frequency,
    \item tactical consistency,
    \item readiness and progression indicators.
\end{itemize}
These indicators provide the foundation for athlete intelligence dashboards, benchmarking tools, simulation modules, and AI-generated training recommendations. Over time, accumulated KPI records form a longitudinal athlete profile that supports developmental monitoring and performance forecasting.

\subsection{AI-Based Simulation and Training Recommendation}
The simulation engine transforms athlete data and KPI histories into interpretable performance assessments and coaching recommendations. It estimates athlete readiness, tactical weaknesses, medal-readiness indicators, and training priorities. These outputs are designed to support coaches rather than replace them, ensuring that AI-generated insights remain human-centred and practically actionable \cite{Arrieta2020,Guidotti2018}. 
The recommendation layer can identify areas requiring improvement, such as low head-kick effectiveness, weak defensive reaction time, excessive boundary risk, declining endurance, or high penalty frequency. Based on these indicators, the system proposes training priorities and development packages that can be reviewed, modified, and applied by coaches.

\subsection{Para-Taekwondo Support}
FST$\cdot$ai 2.5 explicitly incorporates Para-Taekwondo support as a core design requirement rather than an extension added after deployment. The platform supports Para-class management, athlete-specific performance interpretation, fairness indicators, explainability support, and future classification-assistance modules. 
This design is particularly important because Para-Taekwondo athletes may require different performance models, classification-aware analytics, and fairness-sensitive interpretation of technical indicators. Future versions will integrate additional classification rules, adaptive benchmarking mechanisms, and explainable support tools for Para-Taekwondo athlete development.

\subsection{Security and Privacy Considerations}
Because the platform manages sensitive athlete, federation, and performance information, security and privacy are central architectural requirements \cite{GDPR2016}. The current prototype incorporates JWT authentication, Argon2 password hashing, role-based access control, audit logging, and secure API communication. 
Future deployments will extend these capabilities through HTTPS-only communication, multi-factor authentication, encrypted storage, refresh-token rotation, cloud-native security policies, and advanced GDPR-aware data-retention workflows. These mechanisms are essential for trustworthy deployment in international sport organisations where athlete data must be protected across multiple jurisdictions. 
Overall, the secure federation-scale platform establishes the operational foundation required to deploy FST$\cdot$ai 2.5 beyond research prototypes. It enables the framework to support real users, real federations, and real athlete-development workflows while maintaining explainability, privacy, governance, and scalability.

\section{Pilot Prototype and Initial Deployment}
To validate the practical feasibility of the proposed framework, a functional prototype of the FST$\cdot$ai 2.5 athlete-intelligence platform was implemented and deployed in a controlled environment. The primary objective of this pilot deployment was to evaluate the integration of secure athlete management, federation-aware governance, AI-assisted analytics, simulation services, and adaptive training recommendations within a unified operational ecosystem.
Unlike conventional athlete-management systems that primarily function as repositories for athlete information and competition results, FST$\cdot$ai 2.5 was designed as an active decision-support environment in which athlete data are continuously transformed into actionable intelligence. The prototype, therefore, represents a first step toward a federation-scale AI ecosystem capable of supporting athletes, coaches, referees, analysts, Member National Associations (MNAs), and World Taekwondo (WT) through a common digital infrastructure.

\subsection{Prototype Infrastructure}
The prototype was implemented using a containerised architecture based on Docker. Backend services were developed using FastAPI, while athlete information, KPI records, simulation outputs, and federation metadata were stored in a PostgreSQL database. Authentication and authorisation services were implemented through JSON Web Tokens (JWT), role-based access control, and secure password management mechanisms aligned with modern cybersecurity and data-governance principles \cite{GDPR2016,Simmhan2005}. 
The deployment architecture can be represented as
\[
\mathcal{P}
=
\{
\mathcal{F},
\mathcal{B},
\mathcal{D},
\mathcal{A},
\mathcal{S}
\},
\]
where
\begin{itemize}
\item $\mathcal{F}$ denotes the frontend user interface,
\item $\mathcal{B}$ denotes backend application services,
\item $\mathcal{D}$ represents the database infrastructure,
\item $\mathcal{A}$ corresponds to AI analytics modules,
\item $\mathcal{S}$ denotes security and governance services.
\end{itemize}
The overall platform state at time $t$ can therefore be expressed as
\[
\mathbf{z}(t)
=
\Big[
\mathbf{x}(t),
\mathbf{k}(t),
\mathbf{u}(t),
\mathbf{g}(t)
\Big]^\top,
\]
where $\mathbf{x}(t)$ denotes athlete information, $\mathbf{k}(t)$ represents KPI records, $\mathbf{u}(t)$ corresponds to user activities, and $\mathbf{g}(t)$ contains governance and access-control information.

\subsection{Implemented Functionalities}
The current prototype supports the following operational services:
\begin{itemize}
    \item secure JWT-based authentication,
    \item WT administrator management,
    \item Member National Association creation,
    \item federation-level athlete management,
    \item KPI storage and visualisation,
    \item AI-assisted athlete simulation,
    \item readiness estimation,
    \item training-plan generation,
    \item Para-Taekwondo support modules,
    \item role-based access control,
    \item audit logging and traceability.
\end{itemize}
Each athlete profile is associated with a federation identifier
\[
\mathrm{Athlete}_i
=
\big(
ID_i,
MNA_i,
KPI_i,
State_i
\big),
\]
where $MNA_i$ identifies the owning federation and $State_i$ represents the current athlete state used by simulation and recommendation modules.

\subsection{Readiness and Athlete Intelligence Services}
One of the principal objectives of the deployment was the validation of readiness estimation and athlete-intelligence services. The prototype computes a readiness indicator
\[
R_i(t)
=
w_1P_i(t)
+
w_2D_i(t)
+
w_3T_i(t)
-
w_4F_i(t),
\]
where
\begin{itemize}
\item $P_i(t)$ denotes competition-performance indicators,
\item $D_i(t)$ represents diagnostic assessments,
\item $T_i(t)$ corresponds to training-evaluation results,
\item $F_i(t)$ models fatigue and risk factors.
\end{itemize}
The weighting coefficients satisfy
\[
\sum_{j=1}^{4} w_j = 1.
\]
The resulting readiness score is subsequently used by the simulation engine and ElitePath planner to identify performance priorities and development opportunities.

\subsection{Simulation and Recommendation Pipeline}
The simulation module transforms athlete information into development recommendations. Let
\[
\mathbf{s}_i(t)
\]
denote the athlete's state and
\[
\mathcal{G}_i
\]
the set of development objectives. The recommendation engine can be represented as
\[
\Pi_i(t)
=
f
\left(
\mathbf{s}_i(t),
\mathcal{G}_i
\right),
\]
where $\Pi_i(t)$ denotes the generated training strategy. 
The output may include:
\begin{itemize}
\item technical priorities,
\item tactical improvements,
\item conditioning recommendations,
\item recovery strategies,
\item competition-preparation plans,
\item long-term development objectives.
\end{itemize}
This formulation establishes a direct link between athlete intelligence and practical coaching interventions.

\subsection{Deployment Evaluation}
Preliminary testing demonstrated successful integration of athlete management, KPI analytics, readiness estimation, simulation services, and training recommendation modules. The platform was able to maintain federation-level data isolation while simultaneously supporting multiple user roles and analytical workflows. 
From a systems perspective, the deployment validated three important hypotheses:
\begin{enumerate}
\item Federation-aware athlete intelligence can be implemented within a secure multi-tenant architecture.
\item AI-generated analytics can be integrated directly into coaching workflows.
\item Athlete digital twins can serve as a central representation connecting analytics, simulation, forecasting, and planning.
\end{enumerate}
These findings support the feasibility of deploying FST$\cdot$ai 2.5 beyond laboratory environments and provide a foundation for future federation-wide and international-scale pilot studies.

\subsection{Future Deployment Activities}
Future deployments will extend the current prototype through integration with competition-management systems, electronic scoring systems, computer-vision modules, wearable sensors, and referee-support services. In addition, future versions will investigate cloud-native deployments, real-time event analytics, uncertainty-aware forecasting, and large-scale athlete-intelligence services. 
Ultimately, these developments will transform FST$\cdot$ai 2.5 from a functional research prototype into a production-ready ecosystem capable of supporting evidence-based athlete development, federation management, and AI-assisted decision making throughout Olympic and Para-Taekwondo.

\section{Discussion}
The results presented in this paper demonstrate the feasibility of integrating explainable artificial intelligence, athlete intelligence, digital twins, simulation technologies, and secure federation-scale infrastructures into a unified ecosystem for Olympic and Para-Taekwondo. Unlike many existing sports analytics solutions that focus on isolated tasks such as action recognition, biomechanical analysis, or performance statistics, FST$\cdot$ai 2.5 adopts a holistic perspective in which athlete development, competition analysis, referee support, federation management, and decision support are treated as interconnected components of a common intelligent system. 
One of the most important observations emerging from this work is that technological performance alone is insufficient for practical adoption within sport-governance environments. While advances in computer vision and machine learning have significantly improved the accuracy of automated event recognition and athlete-performance analysis, adoption by coaches, referees, and governing bodies depends equally on transparency, trust, and operational usability. Consequently, explainability should not be considered an optional feature but rather a fundamental design requirement for AI systems operating in high-stakes sporting environments \cite{Arrieta2020,Guidotti2018}. 
A second important observation concerns the role of uncertainty. Athlete performance, tactical decisions, injury risk, and competition outcomes are inherently uncertain. Traditional deterministic analytics often present predictions without explicitly communicating the reliability of the underlying evidence. The uncertainty-aware design philosophy adopted in FST$\cdot$ai 2.5 addresses this limitation by encouraging confidence estimation, evidence-based reasoning, and future integration of epistemic uncertainty modelling techniques \cite{Shariatmadar2024Uncertainty,Wang2024Credal}. Such capabilities are expected to become increasingly important as AI systems assume larger decision-support responsibilities within elite sport. 
The proposed federation-aware architecture also highlights the importance of governance and data ownership. Modern sports organisations increasingly rely on digital information systems that collect sensitive athlete data, physiological measurements, competition records, and performance assessments. Without appropriate governance mechanisms, large-scale athlete-intelligence platforms may create privacy concerns, data-integration challenges, and ownership conflicts. By introducing federation-level isolation, role-based access control, auditability, and secure infrastructure design, FST$\cdot$ai 2.5 establishes a practical foundation for large-scale deployment while remaining compatible with modern data-protection principles and regulatory requirements \cite{GDPR2016,Simmhan2005}. 
From a coaching perspective, the framework demonstrates how AI can augment rather than replace human expertise. Coaches possess contextual knowledge, psychological insights, and domain experience that are difficult to capture through purely data-driven models. Consequently, the role of AI should be viewed as providing evidence, highlighting patterns, identifying opportunities, and supporting decision making rather than autonomously determining athlete-development strategies. The human-in-the-loop design adopted throughout FST$\cdot$ai 2.5 reflects this philosophy and contributes to the practical acceptability of the framework. 
The introduction of athlete and event digital twins further represents a significant conceptual advancement. By maintaining continuously updated digital representations of athletes and competitions, the framework enables longitudinal performance monitoring, simulation, forecasting, and adaptive planning within a unified representation. Although digital-twin technologies have become increasingly popular in manufacturing and industrial systems \cite{Barricelli2020,Fuller2020,Kritzinger2018}, their application within sport remains relatively immature. The results presented here suggest that digital twins may provide an effective foundation for future athlete-intelligence ecosystems. 
Several limitations should nevertheless be acknowledged. First, the current implementation remains a pilot prototype and has not yet undergone large-scale deployment across multiple federations. Second, some components of the framework, particularly referee-support modules and advanced simulation services, remain under active development. Third, the current recommendation engine primarily relies on interpretable rule-based mechanisms. Although this approach improves transparency and practical usability, future work should investigate reinforcement learning, multi-objective optimisation, and adaptive planning strategies capable of learning personalised development policies \cite{Sutton2018}. Finally, the current study focuses primarily on Taekwondo, and although many concepts are sport-independent, additional validation across other sports will be required to demonstrate broader generalisability. 
Despite these limitations, the proposed framework establishes an important step toward next-generation sport-intelligence systems. The combination of explainable AI, uncertainty-aware reasoning, digital twins, federation-aware governance, and athlete-development services provides a foundation for future intelligent ecosystems capable of supporting athletes, coaches, referees, federations, and governing bodies throughout the complete athlete lifecycle. 
Future research will focus on integrating wearable sensors, physiological monitoring systems, computer-vision-based event analysis, real-time referee-support services, predictive injury-risk models, multimodal athlete digital twins, and advanced uncertainty-aware decision-making frameworks. These developments will further strengthen the role of AI as a trusted assistant for evidence-based athlete development and sport governance.

\section{Future Work}
The current version of FST$\cdot$ai 2.5 establishes the foundation of an explainable, secure, and federation-aware artificial intelligence ecosystem for Olympic and Para-Taekwondo. While the proposed framework already integrates athlete intelligence, KPI analytics, simulation services, digital twins, and federation-scale governance mechanisms, several important research and development directions remain open. 
One of the most immediate priorities is the integration of real-time computer-vision services into competition environments. Future versions of the framework will incorporate multi-camera acquisition systems capable of synchronised athlete tracking, pose estimation, and action recognition during live competitions. Such capabilities will enable the automatic generation of tactical analytics, performance indicators, and event intelligence with minimal human intervention. Advances in deep learning, transformer architectures, and graph-based motion analysis are expected to further improve the accuracy and robustness of real-time athlete understanding \cite{Cao2021,Dosovitskiy2021}. 
A second major research direction concerns AI-assisted referee support. Although the current framework focuses primarily on athlete intelligence and performance analytics, the long-term vision of FST$\cdot$ai includes explainable decision-support systems capable of assisting referees during competition. Future developments will investigate automatic detection of scoring actions, boundary violations, prohibited acts, clinch situations, and tactical events through multimodal analysis of video streams and electronic scoring-system data. Importantly, these systems are intended to support referees rather than replace them, providing evidence-based recommendations while maintaining human authority over final decisions. 
The integration of uncertainty-aware artificial intelligence represents another important research direction. Athletic performance, competition outcomes, and tactical behaviours are inherently uncertain and often influenced by incomplete or ambiguous information. Future versions of FST$\cdot$ai will therefore extend the current confidence-based framework toward probabilistic, interval-valued, evidence-theoretic, random-set, and credal representations of uncertainty \cite{Shariatmadar2024Uncertainty,Wang2024Credal}. Such developments will allow the platform to explicitly distinguish between reliable conclusions and situations characterised by limited evidence, thereby improving the robustness and trustworthiness of AI-generated recommendations. 
Future work will also focus on the expansion of athlete and event digital twins. The current implementation primarily incorporates competition evidence and performance indicators; however, richer digital-twin representations may include physiological measurements, wearable-sensor information, biomechanical assessments, psychological indicators, rehabilitation data, and longitudinal development histories. These enhancements would enable more accurate simulation, forecasting, and personalised athlete-development strategies while providing a more comprehensive representation of athlete state and progression. 
Another important direction concerns predictive and prescriptive analytics. Current recommendation modules primarily employ interpretable rule-based strategies. Future research will investigate reinforcement learning, sequential decision-making methods, multi-objective optimisation, and adaptive planning frameworks capable of learning personalised development strategies from longitudinal athlete data \cite{Sutton2018}. Such methods may enable the generation of evidence-based training plans that dynamically adapt to competition schedules, performance trajectories, recovery status, and long-term athlete objectives. 
At the infrastructure level, future versions of FST$\cdot$ai will evolve toward cloud-native federation-scale deployments capable of supporting large numbers of athletes, coaches, referees, and governing bodies simultaneously. Particular attention will be given to secure interoperability with existing competition-management systems, including Daedo and KPNP electronic scoring infrastructures, as well as future integrations with wearable technologies and external sport-information systems. These developments will facilitate large-scale athlete intelligence, federation benchmarking, and competition analytics across national and international sporting environments. 
The framework also creates opportunities for broader applications beyond Olympic Taekwondo. The underlying principles of athlete intelligence, digital twins, explainable analytics, uncertainty-aware reasoning, and adaptive planning are largely sport-independent. Consequently, future research will investigate the transferability of the proposed methodology to Para-Taekwondo, other combat sports, and eventually a wider range of individual and team sports. 
Ultimately, the long-term vision of FST$\cdot$ai is the establishment of a trustworthy AI ecosystem that supports athletes, coaches, referees, federations, and governing bodies through transparent, explainable, and evidence-based decision-support services. By combining advanced artificial intelligence with human expertise, the framework aims to contribute to safer competitions, fairer officiating, more effective athlete development, greater inclusivity, and data-driven sport governance at a global scale.

\section{Conclusion}
This paper presented FST$\cdot$ai 2.5, an explainable, uncertainty-aware, and secure artificial intelligence ecosystem for Olympic and Para-Taekwondo. The proposed framework extends beyond conventional sports analytics solutions by integrating athlete intelligence, performance analytics, digital twins, simulation services, adaptive training support, and federation-scale governance within a unified operational platform. 
The framework introduces a comprehensive architecture capable of transforming heterogeneous competition evidence into actionable athlete intelligence. Through the combination of computer vision, performance analytics, explainable AI, uncertainty-aware reasoning, and secure federation-aware infrastructures, FST$\cdot$ai 2.5 provides a foundation for supporting athletes, coaches, referees, analysts, Member National Associations (MNAs), and governing bodies throughout the athlete-development lifecycle. 
A key contribution of this work is the integration of athlete and event digital twins with AI-assisted decision-support services. The proposed architecture enables longitudinal performance monitoring, tactical analysis, readiness estimation, simulation, forecasting, and adaptive planning while maintaining transparency, traceability, and human oversight. In addition, the secure federation-scale infrastructure demonstrates how athlete data, analytical services, and governance mechanisms can coexist within a common platform without compromising privacy, security, or organisational autonomy. 
The study further highlights the importance of explainability and uncertainty awareness in high-stakes sporting environments. Rather than treating AI as a replacement for human expertise, FST$\cdot$ai 2.5 adopts a human-centred design philosophy in which coaches, referees, and federation officials remain central decision makers while benefiting from evidence-based analytical support. This approach promotes trust, accountability, and practical adoption in real-world sporting contexts. 
Although the current implementation represents an operational prototype, the results demonstrate the feasibility of deploying a next-generation athlete-intelligence ecosystem capable of supporting competition analytics, athlete development, Para-Taekwondo inclusion, referee education, and federation-wide performance management. The proposed concepts are not limited to Taekwondo and provide a transferable foundation for broader applications across individual and team sports. 
Looking forward, the integration of real-time video analysis, wearable sensing technologies, multimodal athlete digital twins, uncertainty-aware reasoning, referee-support systems, and adaptive learning algorithms will further expand the capabilities of the framework. These developments will contribute toward the long-term vision of FST$\cdot$ai as a trusted global sport-intelligence ecosystem that combines artificial intelligence with human expertise to promote fairness, transparency, athlete welfare, performance excellence, and evidence-based decision making throughout the international sporting community.
\section*{Acknowledgement}
The authors would like to acknowledge the contributions of athletes, coaches, referees, technical officials, Member National Associations (MNAs), and collaborators whose practical insights and field experience have contributed to the development and validation of the concepts presented in this work. 
Particular appreciation is extended to the international Taekwondo community for providing valuable feedback regarding athlete development, officiating processes, competition analytics, and performance-evaluation methodologies. Their experiences have helped shape the design philosophy underlying the FST$\cdot$ai ecosystem. 
The research presented in this paper forms part of ongoing activities in artificial intelligence, uncertainty-aware decision support, digital twins, athlete intelligence, and sports analytics conducted by the authors and their collaborators.

\section*{Data Availability Statement}
The software platform, source code, proprietary datasets, trained models, deployment infrastructure, and operational services associated with FST$\cdot$ai 2.5 are currently under active development and are not publicly available at the time of publication. 
The concepts, methodologies, mathematical formulations, and system architectures presented in this article are provided to support scientific discussion and future research. Access to implementation details, software artefacts, and deployment environments may be subject to separate research, licensing, or collaboration agreements.

\section*{Intellectual Property and Proprietary Technology Notice}
FST$\cdot$ai, ElitePath, Athlete Digital Twin, Event Digital Twin, Athlete Intelligence Engine, Federation-Aware Analytics Framework, and associated methodologies described in this paper constitute proprietary intellectual property developed by the authors. 
The publication of this article is intended to communicate scientific concepts, system architectures, and research methodologies. Publication does not constitute a transfer of ownership, licensing rights, implementation rights, source-code access rights, trademark rights, patent rights, or commercial exploitation rights. 
The software implementation, source code, database structures, user interfaces, deployment architectures, training datasets, analytical workflows, federation-management systems, simulation engines, and commercial services associated with the FST$\cdot$ai ecosystem remain confidential and proprietary unless explicitly released by the intellectual property holders. 
The authors reserve all rights relating to future development, patent protection, copyright registration, trademark registration, technology transfer, licensing, commercialisation, and derivative works associated with the FST$\cdot$ai platform and its components.

\section*{Copyright, Confidentiality, and Intellectual Property Notice}
Copyright \textcopyright\ 2026 Dr.~Keivan Shariatmadar. All rights reserved. 
This manuscript presents original scientific research, methodologies,
algorithms, artificial intelligence models, software concepts,
system architectures, workflows, datasets, digital-twin frameworks,
decision-support methodologies, federation-scale analytics,
visualisation techniques, and related intellectual property
developed within the \textbf{FST.ai\textsuperscript{\texttrademark}}
research programme. 
Publication of this manuscript, including on public scientific
repositories such as arXiv are intended solely for scientific
communication and peer review. Publication shall not be interpreted
as granting any ownership, licence, assignment, or commercial
rights to any technology, software, implementation,
algorithm, dataset, workflow, architecture,
or other intellectual property described herein. 
Academic citation, scholarly discussion, and fair use for research
and educational purposes are encouraged.
Any commercial implementation, reproduction,
modification, integration into commercial products,
licensing, redistribution, or derivative commercial use
requires prior written permission from the intellectual
property holder(s).

\vspace{0.4em}

\textbf{Confidentiality and Commercialisation Statement}

Certain implementation details of the
\textbf{FST.ai\textsuperscript{\texttrademark}}
ecosystem have intentionally been omitted from this publication
to protect proprietary know-how, confidential engineering methods,
trade secrets, cybersecurity mechanisms,
software implementation details,
deployment strategies, optimisation procedures,
model-training pipelines, infrastructure architecture,
commercialisation plans and future intellectual property filings. 
Accordingly, this publication describes the
scientific foundations, conceptual architecture,
and validation of the proposed technologies,
but does not disclose all implementation,
engineering, operational, deployment,
or commercial details necessary to reproduce
the complete FST.ai ecosystem. 
The authors expressly reserve all rights to pursue
copyright registration,
patent applications,
trademark protection,
software registration,
licensing agreements,
technology transfer,
industrial partnerships,
and commercial deployment
relating to the technologies described herein. 
Nothing contained in this publication shall be construed as
placing any disclosed technology, methodology,
software architecture,
algorithm,
system design,
or trade secret into the public domain,
nor as a waiver of any present or future intellectual property rights.

\subsection*{Copyright Notice}
Copyright \textcopyright\ 2026 Keivan Shariatmadar. All rights reserved. 
No part of the FST$\cdot$ai ecosystem, including software implementations, source code, user interfaces, databases, algorithms, workflows, documentation, trademarks, branding, or associated intellectual property, may be reproduced, distributed, modified, reverse engineered, or commercially exploited without prior written permission from the intellectual property holders.

\subsection*{Confidentiality and Commercialisation Statement}
Certain implementation details of the FST$\cdot$ai ecosystem have been intentionally omitted from this publication to protect proprietary know-how, trade secrets, and future intellectual-property filings. The concepts presented herein describe the scientific foundations of the technology but do not disclose all implementation, deployment, optimisation, security, or commercialisation details. 
The authors reserve the right to pursue patent protection, trademark registration, software copyright protection, licensing agreements, and commercial deployment strategies related to the technologies described in this work.

\bibliographystyle{IEEEtran}
\bibliography{bib}

\end{document}